\documentclass{article}
\usepackage{arxiv}

\usepackage[utf8]{inputenc} 
\usepackage[T1]{fontenc}    
\usepackage{hyperref}       
\usepackage{url}            
\usepackage{booktabs}       
\usepackage{amsfonts}       
\usepackage{nicefrac}       
\usepackage{microtype}      
\usepackage{lipsum}
\usepackage{graphicx}
\graphicspath{ {./images/} }
\usepackage{booktabs}  

\usepackage[round]{natbib} 
\usepackage{xcolor}
\usepackage{caption}
\usepackage{hyperref}
\usepackage{cleveref}
\crefname{figure}{Fig.}{Figs.}
\hypersetup{
    colorlinks=true,
    linkcolor=blue,    
    citecolor=blue,    
    urlcolor=blue      
}

\newcommand{\RN}[1]{\uppercase\expandafter{\romannumeral#1}}

\title{InsurAgent: A Large Language Model-Empowered Agent for Simulating Individual Behavior in Purchasing Flood Insurance }

\author{
Ziheng Geng$^{1,*}$,
Jiachen Liu$^{2,*}$,
Ran Cao$^{3}$,
Lu Cheng$^{4}$,
Dan M. Frangopol$^{5}$,
Minghui Cheng$^{1,6\dagger}$\\
\\
$^{1}$Department of Civil and Architectural Engineering, University of Miami, Coral Gables, FL 33146, USA\\
$^{2}$Department of Electrical and Computer Engineering, University of Miami, Coral Gables, FL 33146, USA\\
$^{3}$College of Civil Engineering, Hunan University, Changsha, 410082, China\\
$^{4}$Department of Computer Science, University of Illinois Chicago, Chicago, IL 60607, USA\\
$^{5}$Department of Civil and Environmental Engineering, Lehigh University, Bethlehem, PA 18015, USA\\
$^{6}$School of Architecture, University of Miami, Coral Gables, FL 33146, USA\\
\\
$^{*}$Equal contribution.\\
$^{\dagger}$Corresponding author: \texttt{minghui.cheng@miami.edu}
}

\begin{document}
\maketitle
\begin{abstract}

Flood insurance is an effective strategy for individuals to mitigate disaster-related losses. However, participation rates among at-risk populations in the United States remain strikingly low. This gap underscores the need to understand and model the behavioral mechanisms underlying insurance decisions. Large language models (LLMs) have recently exhibited human-like intelligence across wide-ranging tasks, offering promising tools for simulating human decision-making. This study constructs a benchmark dataset to capture insurance purchase probabilities across factors. Using this dataset, the capacity of LLMs is evaluated: while LLMs exhibit a qualitative understanding of factors, they fall short in estimating quantitative probabilities. To address this limitation, InsurAgent, an LLM-empowered agent comprising five modules including perception, retrieval, reasoning, action, and memory, is proposed. The retrieval module leverages retrieval-augmented generation (RAG) to ground decisions in empirical survey data, achieving accurate estimation of marginal and bivariate probabilities. The reasoning module leverages LLM common sense to extrapolate beyond survey data, capturing contextual information that is intractable for traditional models. The memory module supports the simulation of temporal decision evolutions, illustrated through a roller coaster life trajectory. Overall, InsurAgent provides a valuable tool for behavioral modeling and policy analysis.

\end{abstract}

\begin{quote}
\end{quote}

\section{Introduction}
Flooding is among the most pervasive and devastating natural hazards worldwide, posing substantial risk to infrastructure reliability, human safety, and socio-economic stability. Under the influence of climate change, the frequency and intensity of extreme precipitation events have markedly increased, contributing to a growing number of flood-related disasters \citep{tezuka2014estimation, liu2020network, cheng2023efficient}. In the United States, Hurricane Harvey in 2017 caused over 100 fatalities and inflicted more than \textdollar{}125 billion in damage \citep{kundis2017harvey, hegar2018storm}. The resulting floods overwhelmed hundreds of thousands of homes and displaced more than 30,000 residents \citep{comes2017hurricane, sebastian2017hurricane}. In response to the growing threat of flood hazards, purchasing flood insurance is widely recognized as an effective strategy for individuals to mitigate associated economic losses. Despite its critical role, participation rate in flood insurance programs remains strikingly low, particularly among high-risk populations. According to a report by the Congressional Budget Office, only 18\% of properties within Special Flood Hazard Areas (SFHAs) were covered by the National Flood Insurance Program (NFIP), with coverage dropping to just 4\% outside these zones \citep{cbo2024flood}. This huge gap between risk exposure and insurance purchase highlights the urgent need to understand the behavioral mechanisms that drive individual decisions regarding flood insurance.

Therefore, extensive research has employed questionnaire surveys to identify the factors that influence individual decisions, given their effectiveness in capturing perceptions and attitudes toward risk \citep{bird2009use, cheng2020investigation, cheng2022life}. These investigations reveal that the decisions to purchase flood insurance result from a complex interplay of economic, geographic, psychological, and socio-demographic factors. Among these factors, objective indicators of economic capacity and flood exposure are consistently found to be key determinants. Households with higher incomes and those residing in higher-valued properties are significantly more likely to purchase flood insurance \citep{browne2000demand, hung2009attitude, shao2017understanding, darlington2022experimental}. Likewise, living within a 100-year floodplain designated by the Federal Emergency Management Agency (FEMA) correlates well with increased insurance uptake \citep{brody2017understanding, netusil2021willingness, RichmondFed2024}. However, objective risk factors cannot translate into action without the consideration of subjective risk perception. Personal experience, particularly recent exposure to flood events, can temporarily elevate perceived risk and increase insurance adoption \citep{petrolia2013risk, lawrence2014integrating, royal2019flood}, though this effect typically fades within a few years as risk salience declines \citep{atreya2015drives, kousky2017disasters}. Additionally, insights from behavior economics explain why many high-risk individuals remain uninsured. Cognitive biases such as short-termism, inertia, and bandwagon effect frequently contribute to suboptimal insurance decision-making \citep{kunreuther2021improving, lacour2025don}. These behavior patterns are not uniform across populations but are shaped by socio-demographic factors, including age, gender, race, and education, which influence how individuals perceive, interpret, and respond to flood risk \citep{atreya2015drives, shao2017understanding, shao2019predicting, eryilmaz2021investigation}. Despite these advancements, existing survey-based models remain limited in scope, covering only a narrow subset of influential features. Moreover, these models rely on an oversimplified representation of complex human behavior as a collection of discrete variables, thereby failing to capture the subtle cognitive processes and context-dependent perceptions underlying real-world decision-making.

Recent advances in large language models (LLMs) have opened new avenues for simulating human behavior across a range of decision-making scenarios. State-of-the-art LLMs, such as ChatGPT \citep{openai2024gpt4o}, Llama \citep{grattafiori2024llama}, DeepSeek \citep{guo2025deepseek}, and Gemini \citep{comanici2025gemini} series, have demonstrated remarkable capabilities in language understanding \citep{karanikolas2023large, min2023recent}, logical reasoning \citep{creswell2022selection, xu2025large}, and in-context learning \citep{liang2025integrating, liu2025large, liang2025automating}. These capabilities have facilitated the development of LLM-empowered agents, which are autonomous entities that integrate modules for planning, memory, and tool use to emulate perception, reasoning, communication, and decision-making in human behavior. Several recent studies have illustrated the application of LLM-empowered agents in modeling human decision-making within domain-specific scenarios. For example, \cite{li2023econagent} proposed EconAgent, which incorporates profile, memory, and action modules to simulate household behavior in labor and consumption decisions. \cite{wang2025user} developed heterogenous agent profiles to replicate user behavior in movie recommender systems, successfully capturing the user conformity phenomenon. \cite{yang2025twinmarket} employed the Belief-Desire-Intention (BDI) framework to structure agent cognitive processes, enabling the simulation of investor behavior in stock market environments. \cite{chen2023put} designed agents for sequential decision-making in dynamic auction settings, exhibiting skills such as budget management, strategic planning, and goal adherence. \cite{xu2025assessing} developed agents to simulate household interpretations of pre-disaster warning information and assess their underestimation of hurricane food shortages. \cite{chen2025perceptions} introduced a behavioral theory-informed agent to simulate individual decisions during wildfire evacuation. These agents serve as digital sandboxes for simulating, observing, and analyzing individual and collective behaviors within a society.

This study investigates the potential of LLMs in simulating individual flood insurance decision-making. The Llama-3.3 70B model is selected for its open-source accessibility and deployment scalability. To facilitate performance assessment, Section 2 constructs a benchmark dataset based on \cite{shao2017understanding} to capture marginal and bivariate probability distributions of insurance purchase across key features. Using this dataset, Section 3 evaluates LLM’s common sense, revealing a qualitative grasp of factors but an inability to produce accurate quantitative estimates. To address this limitation, Section 4 proposes InsurAgent, an LLM-empowered agent to emulate individual insurance decisions. The agent comprises five modules: perception, retrieval, reasoning, action, and memory. Specifically, the perception module interprets user profiles and extracts relevant information. The retrieval module employs retrieval-augmented generation (RAG) to ground probability estimations in empirical survey data. The reasoning module leverages LLM’s common sense to extrapolate beyond survey data. The action module integrates the reasoning results to generate purchase probabilities. The memory module supports sequential and temporal decision modeling. Experimental results in Section 5 show that InsurAgent closely aligns with benchmark probabilities for both marginal and bivariate probability estimations, consistently outperforming state-of-the-art LLMs. It also demonstrates strong extrapolation capabilities, capturing contextual information such as resident cities, social environment, and prior experience. The effectiveness of the memory module is validated through a simulated “three-up, three-down” life trajectory.

\section{Dataset Preparation}
\label{sec:headings}

To assess the performance of LLM in simulating individual flood insurance decisions, a benchmark dataset is constructed based on the regression model reported in \cite{shao2017understanding}. The original study analyzed survey responses from U.S. Gulf Coast residents to identify factors influencing the voluntary flood insurance purchase. These factors span four categories: sociodemographic characteristics, home ownership, distance from the coast, and perceived flood-related risk. Particularly, sociodemographic variables include age, gender, education, and income. Each variable was discretized into categorical levels and then a mixed-effects logit regression model was fitted to quantify their effects. The results reveal four tiers of statistical significance: high (\(p < 0.001\)) for education, income, and distance from the coast; medium (\(0.001 \le p < 0.01\)) for home ownership; low (\(0.01 \le p < 0.05\)) for perceived flooding amount and perceived hurricane strength; and minimal (\(p \ge 0.05\)) for age, gender, perceived hurricane number, and belief in climate change, where \emph{\(p\)} denotes the \emph{p}-value used in hypothesis testing. For more details, readers are referred to \cite{shao2017understanding}.

\setlength{\aboverulesep}{0.25ex}  
\setlength{\belowrulesep}{0.25ex} 
\begin{table}[htbp]
\centering
\caption{Marginal probabilities of purchasing insurance by individual-level features (derived from Shao et al. 2017).}
\label{table1}
\begin{tabular}{llc llc}
\toprule
\textnormal{\textbf{Individual-level features}} & \textnormal{\textbf{Code}} & \textnormal{\textbf{Marginal probability}} & \textnormal{\textbf{Individual-level features}} & \textnormal{\textbf{Code}} & \textnormal{\textbf{Marginal probability}} \\
\midrule
\multicolumn{3}{l}{\textbf{Socio-demographic}} & \multicolumn{3}{l}{\textbf{Home Ownership**}} \\
\multicolumn{3}{l}{\textbf{Age}} & \quad Own   & - & 0.269 \\
\quad18--24      & 1 & 0.233 & \quad Rent  & - & 0.113 \\\cmidrule(lr){4-6}
\quad25--34      & 2 & 0.238 & \multicolumn{3}{l}{\textbf{Distance from the Coast***}} \\
\quad35--44      & 3 & 0.242 & \quad On the water         & 1 & 0.394 \\
\quad45--54      & 4 & 0.247 & \quad Near the water       & 2 & 0.340 \\
\quad55--64      & 5 & 0.251 & \quad Within 2-5 miles    & 3 & 0.290 \\
\quad65+         & 6 & 0.256 & \quad 5-10 miles          & 4 & 0.242\\\cmidrule(lr){1-3}
\multicolumn{3}{l}{\textbf{Gender}} & \quad 11-30 miles        & 5 & 0.201 \\
\quad Female     & 1 & 0.242 & \quad 31-60 miles         & 6 & 0.165  \\
\quad Male       & 0 & 0.262 & \quad > 60 miles          & 7 & 0.133 \\ 
\cmidrule(lr){1-3}
\cmidrule(lr){4-6}
\multicolumn{3}{l}{\textbf{Education***}}   & \multicolumn{3}{l}{\textbf{Risk Perceptions}} \\ 
\quad Less than high school         & 1 & 0.168 &  \multicolumn{3}{l}{\textbf{Flood Amount*}}\\
\quad High school degree            & 2 & 0.203 & \quad Decreased            & -1 & 0.220 \\
\quad Some college         & 3 & 0.244 & \quad About the same  & 0 & 0.248 \\
\quad College degrees      & 4 & 0.288 & \quad Increased   & 1 & 0.278  \\ 
\cmidrule(lr){1-3} 
\cmidrule(lr){4-6}
\multicolumn{3}{l}{\textbf{Income***}}     & \multicolumn{3}{l}{\textbf{Hurricane Number}}  \\
\quad Under \$10,000    & 1 & 0.128 & \quad Decreased          & -1 & 0.236 \\
\quad \$10--19,999      & 2 & 0.149 & \quad About the same     & 0 & 0.250 \\
\quad \$20--29,999      & 3 & 0.174 & \quad Increased          & 1 & 0.265 \\ \cmidrule(lr){4-6}
\quad \$30--39,999      & 4 & 0.201 & \multicolumn{3}{l}{\textbf{Hurricane Strength*}} \\
\quad \$40--49,999      & 5 & 0.232 & \quad Decreased          & -1 & 0.212 \\
\quad \$50--74,999      & 6 & 0.265 & \quad About the same     & 0 & 0.241 \\
\quad \$75--99,999      & 7 & 0.301 & \quad Increased    & 1 & 0.273 \\
\cmidrule(lr){4-6}
\quad \$100,000+        & 8 & 0.340 & \multicolumn{3}{l}{\textbf{Belief in Climate Change}} \\ 
 & & & \quad Happening           & 1 & 0.255 \\
 & & & \quad Not happening       & 0 & 0.233 \\
 \bottomrule
\multicolumn{6}{l}{\footnotesize Statistical significance: *** High, ** Medium, * Low, and Minimal.}
\end{tabular} 
\end{table}

While the original study reports the feature distributions of the surveyed population and the fitted mixed-effects logit regression model, it does not provide insurance purchase probabilities by individual-level features. To address this gap, a Monte Carlo simulation is performed to reconstruct these probabilities. Specifically, a synthetic population of 10 million individuals is first generated, with each individual’s attributes independently sampled from the reported feature distributions. These profiles are then input into the mixed-effects logit regression model to compute the log-odds of purchasing insurance, which are converted to purchase probabilities using the logistic sigmoid function. To simulate binary purchase behavior (purchase or no-purchase), Bernoulli trials are applied to each computed probability. By aligning the simulated population-level purchase rate with the survey-reported value, this procedure effectively reproduces the insurance purchase decisions of the population. 

To quantify the impact of specific features on insurance purchase decisions, the simulated population and purchase outcomes are analyzed in terms of marginal and bivariate probability distributions. The marginal probability distribution captures the influence of individual factors, where the purchase probability for each categorical level of a given variable is computed as the proportion of individuals with that attribute who purchased insurance. This reflects the marginal effect of each feature on insurance decisions, as summarized in Table~\ref{table1}. In addition, bivariate probability distributions are constructed for pairs of variables with differing levels of statistical significance (e.g., high vs. medium, high vs. low, medium vs. minimal). For each pair, all combinations of categorical levels are enumerated, and the corresponding bivariate purchase probabilities are computed. Together, the marginal and bivariate probabilities establish a comprehensive benchmark for evaluating the ability of LLMs to simulate human decision-making in flood insurance purchase.

\section{Evaluation of LLM’s Capability in Flood Insurance Decision-Making}
\label{sec:others}

Using the benchmark dataset, this section evaluates the capability of LLMs to simulate individual decision-making in purchasing flood insurance. The assessment is conducted from two aspects: qualitative and quantitative. For the qualitative assessment, a third-person discriminative task is designed to test whether the LLM demonstrates a common sense of influencing factors. This is achieved by instructing LLM to rank the purchase probabilities of various individual profiles. For the quantitative assessment, a first-person role-playing task is designed where the LLM is prompted to adopt the role of a specific individual with a given background and to estimate the probability of purchasing insurance. Collectively, these two tasks enable a systematic assessment of the LLM’s capacity to model individual flood insurance decisions.

\begin{figure}[ht!]
\centering
\includegraphics[width=0.9\textwidth]{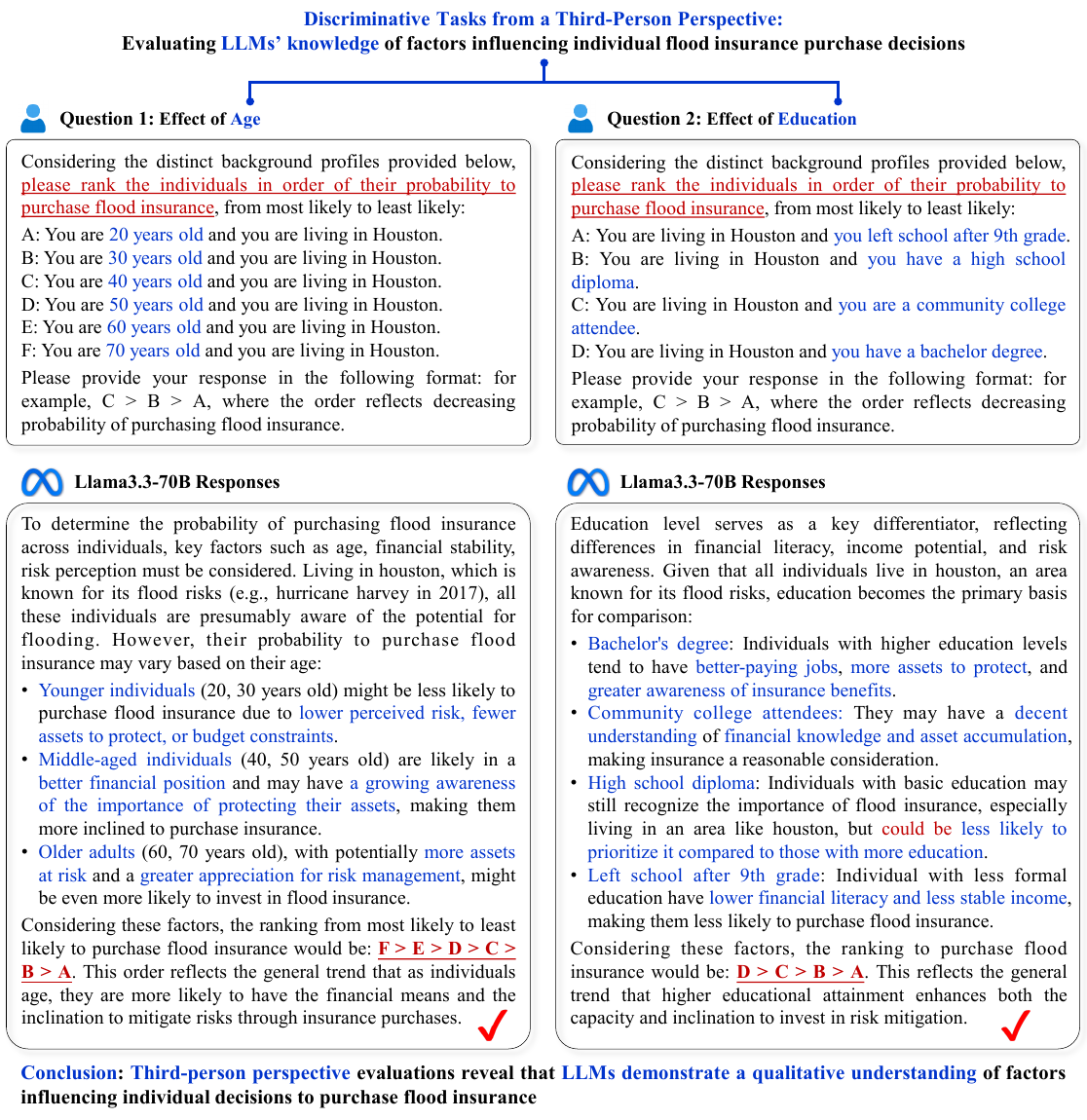}
\caption{Illustration of correct responses of LLM in the third-person discriminative tasks: LLM demonstrates a qualitative understanding of factors influencing individual flood insurance decisions.}
\label{Figure1}
\end{figure}

\begin{figure*}[htbp]
\centering
\includegraphics[width=0.76\textwidth]{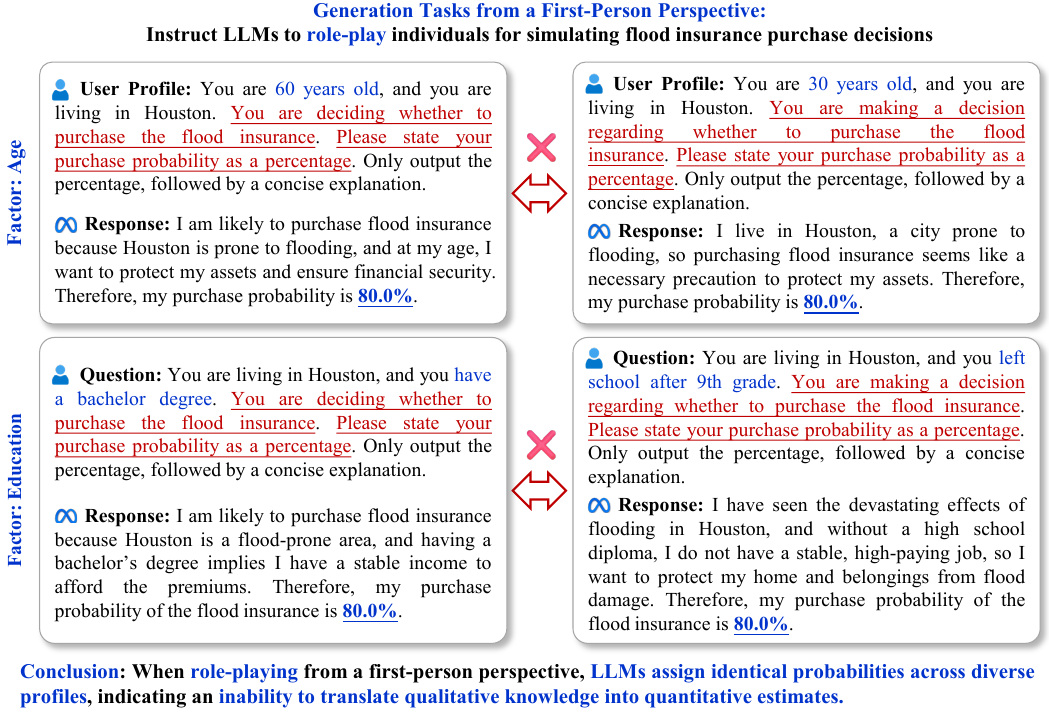}
\caption{Illustration of LLM failure in first-person role-playing tasks: LLMs fail to capture the differences in insurance purchase probabilities across individual profiles.}
\label{Figure2}
\end{figure*}

\setlength{\aboverulesep}{0.25ex}  
\setlength{\belowrulesep}{0.25ex} 
\begin{table}[htbp]
\centering
\caption{Marginal probabilities of purchasing flood insurance estimated by Llama-3.3 70B model.}
\label{table2}
\begin{tabular}{l c@{\hspace{2.5em}} c  l  c@{\hspace{2.5em}} c}
\toprule
\multicolumn{1}{c}{\textnormal{\textbf{Individual-level features}}} &
\multicolumn{2}{c}{\textnormal{\textbf{Marginal probability}}} &
\multicolumn{1}{c}{\textnormal{\textbf{Individual-level features}}} &
\multicolumn{2}{c}{\textnormal{\textbf{Marginal probability}}} \\
 & \textnormal{Estimated} & \textnormal{Benchmark} & & \textnormal{Estimated} & \textnormal{Benchmark} \\
\midrule
\multicolumn{3}{l}{\textbf{Socio-demographic}} & \multicolumn{3}{l}{\textbf{Home Ownership**}} \\
\multicolumn{3}{l}{\textbf{Age}} & \quad Own & 0.850 & 0.269 \\
\quad18--24      & 0.800 & 0.233 & \quad Rent & 0.815 & 0.113 \\\cmidrule(lr){4-6}
\quad25--34      & 0.800 & 0.238 & \multicolumn{3}{l}{\textbf{Distance from the Coast***}} \\
\quad35--44      & 0.800 & 0.242 & \quad On the water         & 0.900 & 0.394 \\
\quad45--54      & 0.800 & 0.247 & \quad Near the water       & 0.900 & 0.340 \\
\quad55--64      & 0.800 & 0.251 & \quad Within 2--5 miles     & 0.900 & 0.290 \\
\quad65+         & 0.830 & 0.256 & \quad 5--10 miles           & 0.800 & 0.242 \\\cmidrule(lr){1-3}
\multicolumn{3}{l}{\textbf{Gender}} & \quad 11--30 miles        & 0.800 & 0.201 \\
\quad Female     & 0.800 & 0.242 & \quad 31--60 miles          & 0.800 & 0.165 \\
\quad Male       & 0.800 & 0.262 & \quad > 60 miles            & 0.800 & 0.133 \\\cmidrule(lr){1-3}
\cmidrule(lr){4-6}
\multicolumn{3}{l}{\textbf{Education***}} & \multicolumn{3}{l}{\textbf{Risk Perceptions}} \\
\quad Less than high school         & 0.800 & 0.168 & \multicolumn{3}{l}{\textbf{Flood Amount*}}\\
\quad High school degree            & 0.800 & 0.203 & \quad Decreased            & 0.200 & 0.220 \\
\quad Some college                  & 0.800 & 0.244 & \quad About the same       & 0.200 & 0.248 \\
\quad College degrees               & 0.800 & 0.288 & \quad Increased            & 0.900 & 0.278 \\\cmidrule(lr){1-3}
\cmidrule(lr){4-6}
\multicolumn{3}{l}{\textbf{Income***}} & \multicolumn{3}{l}{\textbf{Hurricane Number}}  \\
\quad Under \$10,000    & 0.800 & 0.128 & \quad Decreased          & 0.200 & 0.236 \\
\quad \$10--19,999      & 0.800 & 0.149 & \quad About the same     & 0.800 & 0.250 \\
\quad \$20--29,999      & 0.800 & 0.174 & \quad Increased          & 0.900 & 0.265 \\\cmidrule(lr){4-6}
\quad \$30--39,999      & 0.800 & 0.201 & \multicolumn{3}{l}{\textbf{Hurricane Strength*}} \\
\quad \$40--49,999      & 0.800 & 0.232 & \quad Decreased          & 0.200 & 0.212 \\
\quad \$50--74,999      & 0.800 & 0.265 & \quad About the same     & 0.800 & 0.241 \\
\quad \$75--99,999      & 0.800 & 0.301 & \quad Increased          & 0.900 & 0.273 \\
\cmidrule(lr){4-6}
\quad \$100,000+        & 0.800 & 0.340 & \multicolumn{3}{l}{\textbf{Belief in Climate Change}} \\ 
 & & & \quad Happening           & 0.900 & 0.255 \\
 & & & \quad Not happening       & 0.200 & 0.233 \\
\bottomrule
\multicolumn{6}{l}{\footnotesize Statistical significance: *** High, ** Medium, * Low, and Minimal.}
\end{tabular}
\end{table}

\subsection{Qualitative assessment of factors }

A series of discriminative tasks are constructed to evaluate whether LLMs possess qualitative knowledge of factors influencing individual flood insurance decisions. Each task isolates a single factor by presenting multiple individual profiles that are identical in all attributes except the factor under evaluation. The LLM is prompted to rank these profiles in descending order of their probability to purchase flood insurance from a third-person perspective. This experiment directly assesses the LLM's understanding of key decision variables, providing qualitative insights into its capacity to simulate human behavior in purchasing flood insurance.

The results from the third-person discriminative tasks indicate that the Llama-3.3 70B model exhibits a strong qualitative understanding of all influencing factors. \cref{Figure1} presents two illustrative examples of correct responses, in which the LLM evaluates the impact of age and education. In both cases, the LLM not only produces accurate rankings of individuals’ probability to purchase flood insurance but also provides sound reasoning for its conclusions. For age, the LLM correctly recognizes that older individuals are more likely to purchase flood insurance due to greater asset accumulation and heightened risk awareness. Similarly, for education, the LLM infers that individuals with higher educational attainment demonstrate stronger financial literacy and a better understanding of insurance benefits, leading to a higher propensity to purchase flood insurance. These results suggest that the LLM effectively identifies key factors driving individual differences in decision-making and captures the impact of these factors. This confirms that Llama-3.3 70B model has qualitative knowledge required to understand flood insurance purchase decisions.

\subsection{Quantitative estimation of probabilities}

The preceding subsection has confirmed that the Llama-3.3 70B model possesses a qualitative understanding of factors influencing flood insurance decisions. Building on this finding, this subsection aims to answer a more challenging question: can LLM leverage these knowledge to simulate human behavior from a quantitative perspective? To explore this, a first-person role-playing task is designed in which the LLM is assigned a specific user profile and prompted to assume that role in deciding whether to purchase flood insurance. The LLM is instructed to output a purchase probability, followed by a concise explanation. Unlike the third-person discriminative tasks, which involve comparative evaluation across multiple individual profiles, the role-playing task requires the LLM to generate a context-specific probability prediction based on a single user profile. Success in this task depends on the LLM’s ability to activate domain knowledge and translate qualitative understanding into quantitative probabilities.

The results from the first-person role-playing tasks reveal a significant failure of Llama-3.3 70B model in simulating human decision-making. For each individual profile, the LLM is prompted 20 times to estimate the probability of purchasing flood insurance. The average results are reported in Table~\ref{table2}, and two key patterns emerge. First, LLM exhibits a strong risk-averse bias, consistently generating high purchase probabilities, typically at or above 0.800. Second, the LLM demonstrates limited sensitivity to variations in individual-level features. For factors such as gender, education, and income, LLM fails to differentiate between subgroups, producing identical outputs across distinct profiles. For other factors including age, home ownership, distance from the coast, and risk perception, the LLM correctly identifies the impact of factors but fails to adjust the probability magnitude accordingly. \cref{Figure2} illustrates this failure using examples of age and education. It shows that while LLM generates contextually appropriate reasoning, such as individuals with higher education are more likely to have stable income, it consistently assigns the same purchase probability of 0.800 across cases. This highlights a critical knowledge-to-action gap: while the LLM demonstrates a qualitative understanding of factors influencing flood insurance decisions, it fails to translate this understanding into quantitative probability estimations. Therefore, this paper attempts to leverage the qualitative understanding to develop LLM-empowered agents that can accurately estimate probabilities.

\section{InsurAgent: An LLM-empowered Agent for Insurance Decision Simulation}
\label{sec:others}

To address the knowledge-to-action gap observed in role-playing tasks, an LLM-empowered agent, InsurAgent, is proposed to simulate individual decision-making in flood insurance purchase. The core design philosophy of InsurAgent draws on the heuristic decision-making paradigm, which indicates that individuals often consult empirical data or the experiences of similar others to inform their decisions, particularly in scenarios involving risk and uncertainty \citep{simon1955behavioral, payne1992behavioral}. Inspired by this cognitive pattern, InsurAgent integrates region-specific survey data that provide population-level purchase probabilities associated with key factors (Table~\ref{table1}). These data serve as empirical baseline, enabling the agent to calibrate its understanding of risk and probability. By aligning the agent’s reasoning with real-world statistics, InsurAgent is expected to produce accurate, personalized, and context-specific simulations of flood insurance decisions.

\subsection{Agent architecture}

\begin{figure}[!ht]
\centering
\includegraphics[width=0.8\textwidth]{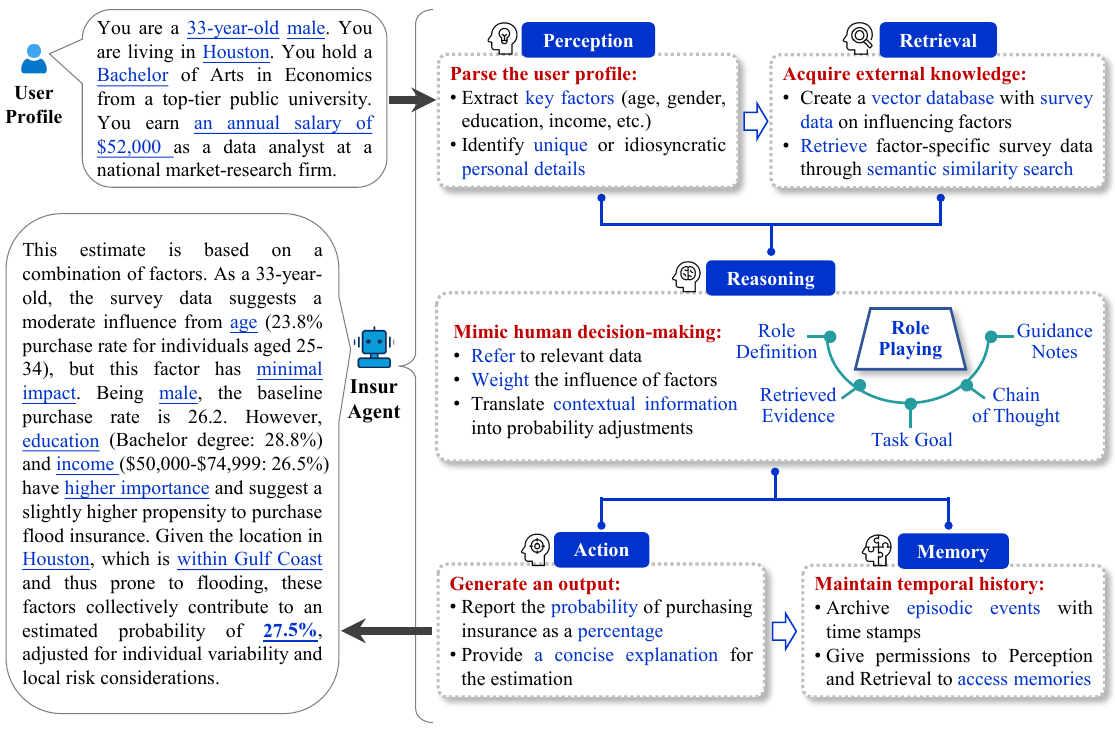}
\caption{Architecture of the InsurAgent, comprising five modules: perception, retrieval, reasoning, action, and memory.}
\label{Figure3}
\end{figure}

The InsurAgent is designed with five core modules: perception, retrieval, reasoning, action, and memory, as illustrated in \cref{Figure3}. This modular design enables a structured, multi-stage decision-making that mimics human cognition process. The procedure begins with the perception module, which parses the input user profile and extracts two types of information: (a) factors that correspond to variables presented in the survey data, and (b) unique personal details not explicitly captured by predefined variables. This architecture ensures that both statistical indicators and contextual information are available to inform downstream decision-making. The extracted factors are then passed to the retrieval module, which interfaces with a region-specific survey database. Through semantic similarity search, the retrieval module identifies factor-specific statistics that reflect population-level flood insurance purchase probabilities. These empirical data serve as reference points to guide the LLM’s reasoning.

The retrieved survey data, along with the parsed output from the perception module, are fed into the reasoning module, which serves as the cognitive core of the InsurAgent. Herein, the agent performs a first-person role-playing task to emulate human decision-making. The reasoning process unfolds in three stages. First, the agent organizes all retrieved survey data and treats them as quantitative benchmarks. Second, it weighs the influence of each factor to derive a baseline purchase probability. Third, the agent applies common sense to extrapolate the baseline probability based on unique personal details, translating contextual information into quantitative probability adjustments. The reasoning process of the InsurAgent is informed by a structured prompt template, which will be elaborated in the next subsection.

Following the reasoning process, the action module generates a final prediction of flood insurance purchase probability, reported as a percentage between 0\% and 100\%. This estimation is accompanied by a concise explanation that summarizes the rationale behind the decision, thereby improving the transparency and interpretability. To support sequential decision tasks and cumulative behavioral modeling, the memory module is designed to archive the reasoning trajectory and final probability output with temporal stamps, forming an episodic history. These memory traces can be queried by the perception and retrieval modules in future interactions, allowing the agent to simulate evolving decisions in response to life events or environment changes. An example of the functionality of the memory module is presented in Section 5.4. Together, these five modules constitute the cognitive architecture of InsurAgent, enabling it to simulate individual flood insurance decisions.

\subsection{Prompt design}

\begin{figure*}[!ht]
\centering
\includegraphics[width=0.8\textwidth]{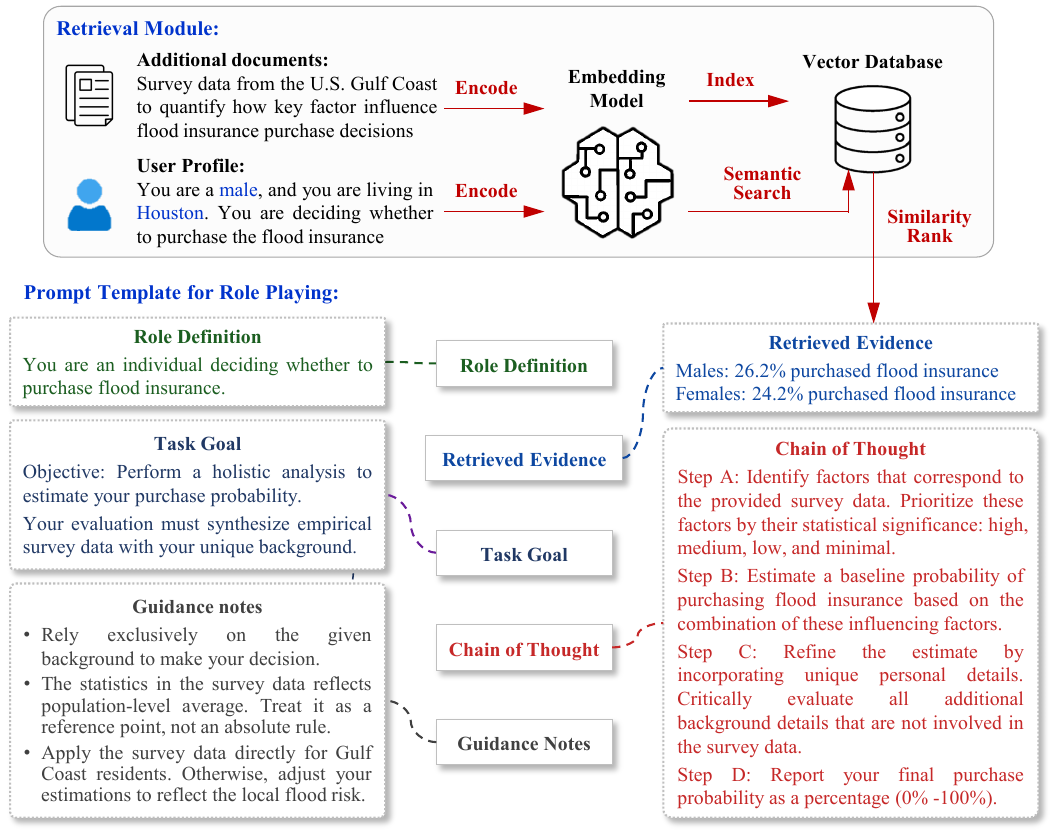}
\caption{Prompt template used by InsurAgent for role-playing tasks, where empirical survey data on factors are retrieved from an external database and integrated into the system prompt.}
\label{Figure4}
\end{figure*}

To support the structured reasoning of InsurAgent, a prompt template is proposed, as demonstrated in \cref{Figure4}. This template scaffolds the agent’s step-by-step decision-making process, guiding it toward producing realistic and context-specific behaviors. It consists of five components: role definition, retrieved evidence, task goal, chain of thought, and guidance notes, each of which plays a distinct role in shaping the agent’s generative behavior. Specifically, the role definition component instantiates the agent’s persona by instructing it to role-play an individual deciding whether to purchase flood insurance. This activates the LLM’s domain knowledge of flood-related decision-making and ensures a consistent first-person narrative throughout the response.

The retrieved evidence component is dynamically populated based on the user profile. As described in the previous section, this retrieve process acquires relevant statistics from survey data collected in the U.S. Gulf Coast region. Specifically, the survey data used here comprise population-level flood insurance purchase probabilities associated with 10 factors, as listed in Table~\ref{table1}. The detailed procedures for implementing RAG are as follows. Survey documents are first encoded into embeddings and then factor-specific statistics are indexed and stored as a vector database. When a user profile is presented, the perception module extracts relevant factors, which are then embedded and semantically matched against the database to retrieve corresponding statistics. These retrieved benchmarks are inserted into the prompt as external references to inform reasoning. The task goal component illustrates the primary objective: to perform a holistic analysis that synthesizes empirical data with individual background to estimate the probability of purchasing flood insurance.

At the core of the agent’s cognitive workflow is the chain of thought (CoT) component, which guides reasoning process through four stages, as demonstrated in \cref{Figure4}. (a) Factor identification and prioritization: The agent identifies relevant factors from the user profile and prioritizes them based on statistical significance levels, as established in Section 2. (b) Baseline estimation: Using the retrieved evidence for these factors, the agent estimates a baseline purchase probability by assessing their relative influence. (c) Adjustment according to contextual information: The agent applies its common sense to interpret personal details, adjusting the baseline probability upwards or downwards. (d) Final integration: The agent integrates the baseline and adjustments to generate the final purchase probability as a percentage.

The guidance notes component imposes constraints on the agent’s generative behavior. First, it instructs the agent to rely exclusively on the provided background to reduce hallucination and enhance reproducibility. Second, it emphasizes that the retrieved statistics represent population-level averages used as reference points rather than absolute rules, allowing the agent to flexibly adjust these empirical benchmarks based on the specific user profile. Third, it specifies that the empirical benchmarks are geographically specific. For users located outside the Gulf Coast region, the agent should apply domain knowledge to adjust probability estimates in accordance with local flood risk. Collectively, these five components transform the LLM from a general-purpose text generator into a specialized agent capable of performing complex behavior simulations.

\begin{figure*}[!ht]
\centering
\includegraphics[width=0.8\textwidth]{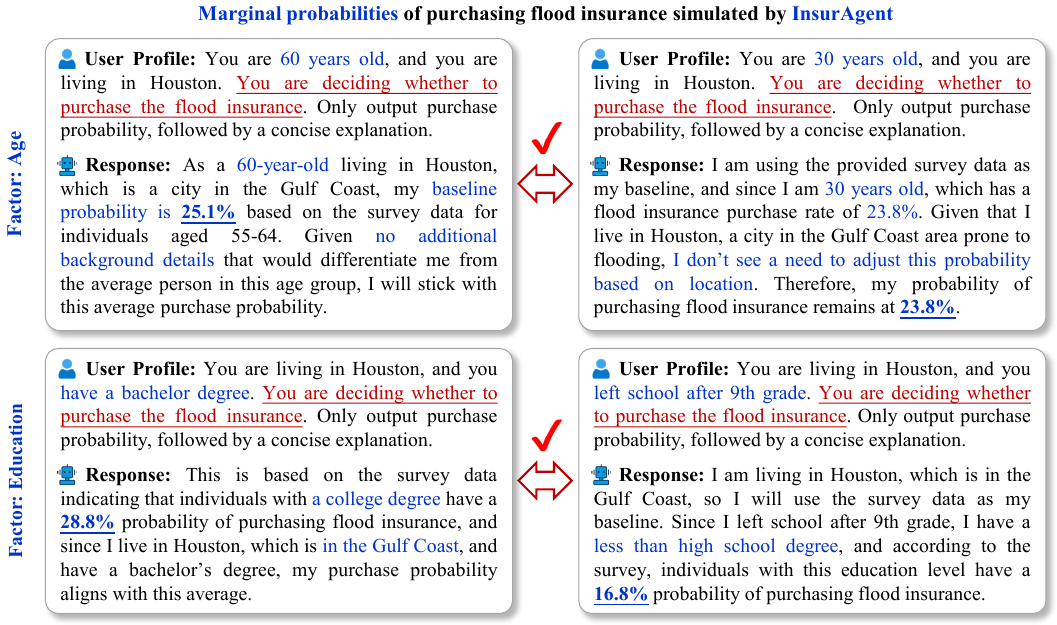}
\caption{Representative examples of InsurAgent-simulated marginal probabilities of purchasing flood insurance.}
\label{Figure5}
\end{figure*}

\section{Results and discussion}

\subsection{Performance of InsurAgent in estimating marginal and bivariate probabilities}

The performance of the proposed InsurAgent is assessed by comparing the estimated flood insurance purchase probability against benchmark dataset across marginal and bivariate distributions. These scenarios are designed to evaluate the agent’s ability to retrieve relevant data, follow quantitative references, and estimate the probabilities under combined effects. For marginal probability estimations, each user profile contains only one variable. These cases assess the agent’s capacity to align its predictions with empirical references, thereby minimizing the risk of hallucinations. For bivariate probability estimations, user profiles are constructed to include two variables, which are selected to span varying levels of statistical significance. These scenarios evaluate the agent’s capacity to synthesize multiple empirical statistics to produce coherent predictions.

For marginal probability estimations, InsurAgent accurately predicts the empirical purchase probabilities retrieved from the external survey database across all factors. Specifically, each user profile in this setting includes a single variable along with a specified resident city within the U.S. Gulf Coast region. With no additional contextual information, the agent is expected to retrieve the corresponding population-level statistics as the predicted probability. \cref{Figure5} illustrates InsurAgent’s responses under two representative scenarios. In the case of age, the agent correctly retrieves baseline probabilities for individuals aged 60 (25.1\%) and 30 (23.8\%). It explicitly states that, given no additional background details, it would adhere to the average probability for the specified age group. Similarly, for the education factor, the agent accurately retrieves the empirical probabilities for individuals with a bachelor’s degree (28.8\%) versus those who left school after 9th grade (16.8\%). This behavior is cognitively plausible, mirroring how humans rely on average trends to inform decisions in the absence of additional contextual information. These results affirm the effectiveness of InsurAgent’s architecture: the retrieval module identifies the corresponding statistical data, and the reasoning module adheres to prompt instructions to use this data as the decision basis.

\begin{figure*}[htbp]
\centering
\includegraphics[width=0.4\textwidth]{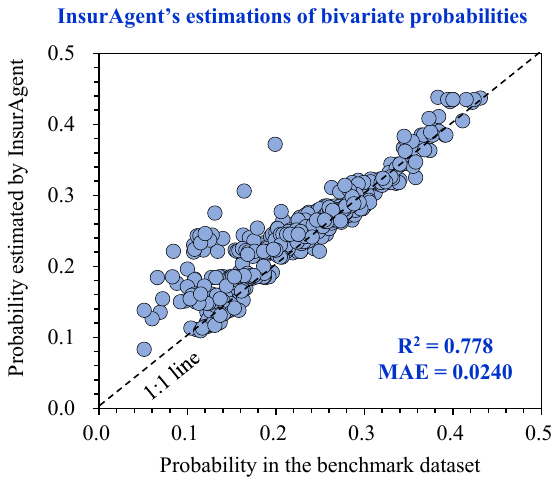}
\caption{Comparison of bivariate probabilities estimated by InsurAgent and benchmark dataset.}
\label{Figure6}
\end{figure*}

The InsurAgent also exhibits strong alignment with benchmark dataset in estimating bivariate probabilities. In this setting, each user profile has two variables along with a residential location within the U.S. Gulf Coast region. Since the external database contains only marginal probabilities, this task requires the agent to integrate separate empirical references and infer their combined effects through extrapolation. Note that the two variables are intentionally selected from different levels of statistical significance, such as education (high) and perceived flooding amount (low). The results show that the agent’s estimated purchase probabilities closely match those in the benchmark dataset, yielding a correlation of determination (R\textsuperscript{2}) of 0.778 and a mean absolute error (MAE) of 0.0240, as depicted in \cref{Figure6}. The high R\textsuperscript{2} and low MAE values demonstrate that InsurAgent effectively integrates multiple empirical evidence through its structured reasoning process.

\begin{figure}[!ht]
\centering
\includegraphics[width=0.8\textwidth]{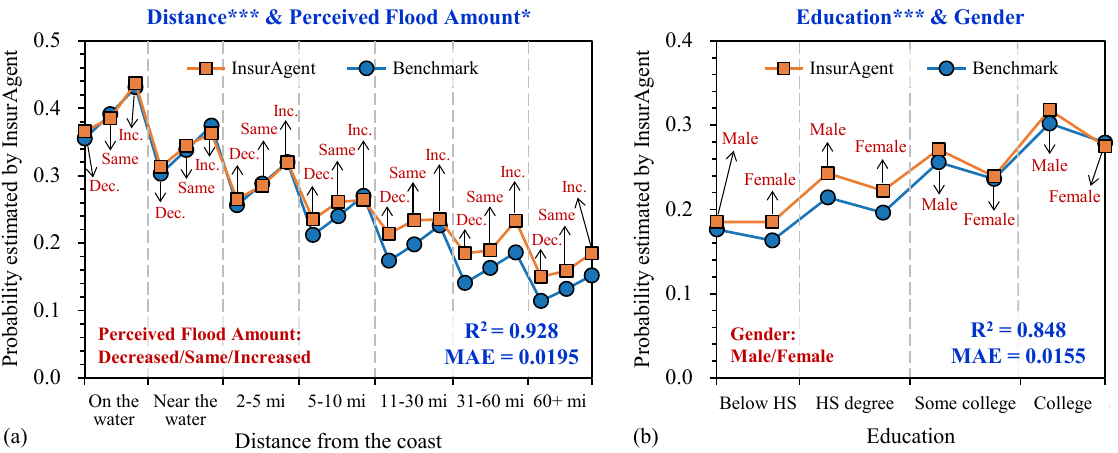}
\caption{InsurAgent-simulated bivariate probabilities of purchasing flood insurance: (a) distance from the coast and perception of flood amount, and (b) education and gender.}
\label{Figure7}
\end{figure}

In addition to producing accurate point estimates, the InsurAgent effectively captures the trend patterns consistent with the benchmark, as shown in \cref{Figure7}. These representative scenarios involve combinations of variables with varying levels of statistical significance. Across both cases, the agent’s predicted trends closely align with those in the benchmark dataset. The agent first establishes a baseline from the dominant variable and then refines the estimate using the secondary variable. For instance, in the scenario combining distance from the coast with perceived flood amount, major shifts are driven by changes in distance. Similarly, in scenarios involving education and gender, the primary variable shapes the overall trajectory, while the secondary variable introduces adjustments. This behavior underscores the agent’s capacity to execute chain-of-thought reasoning to integrate multiple pieces of evidence.

\begin{figure*}[!ht]
\centering
\includegraphics[width=0.8\textwidth]{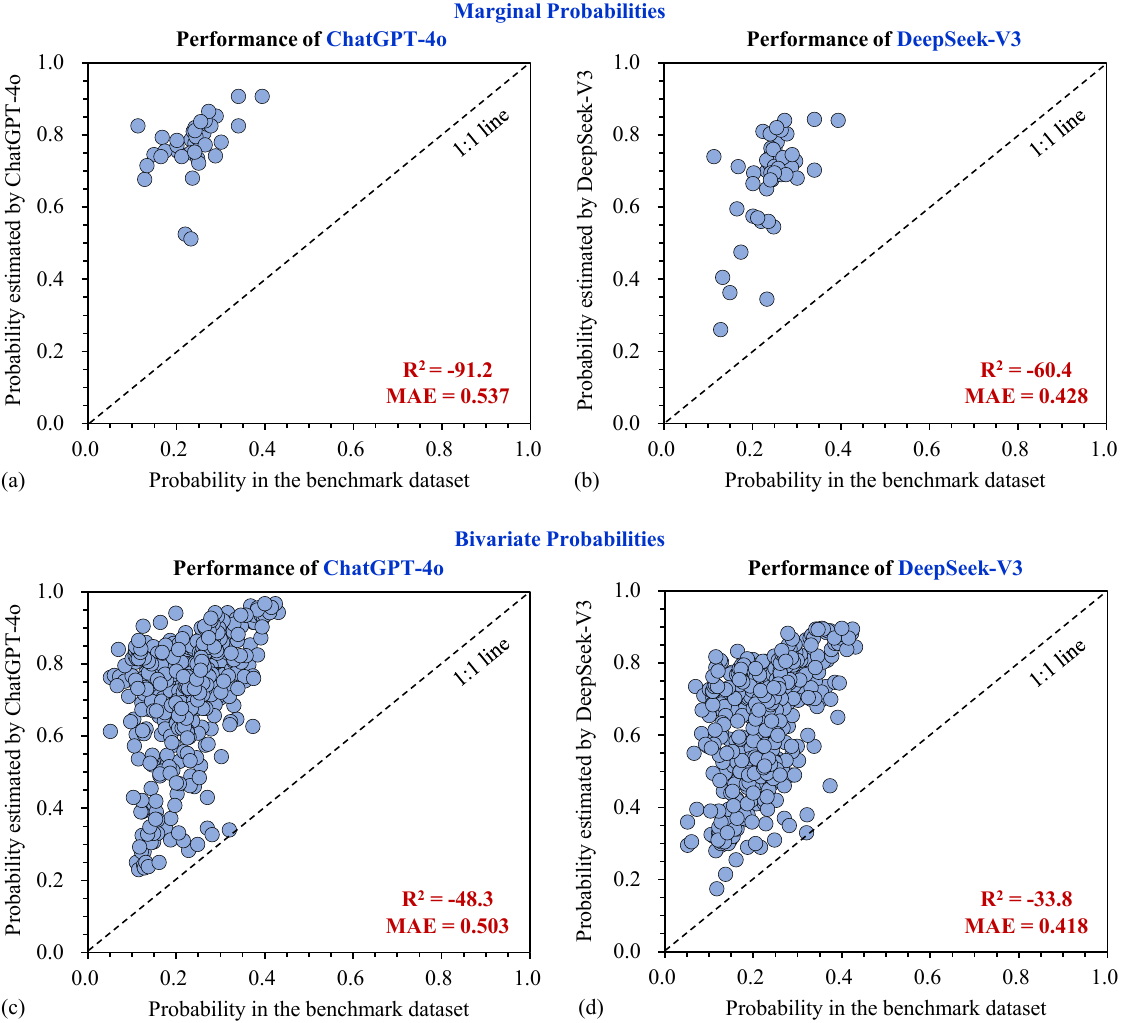}
\caption{Comparison of marginal and bivariate probabilities estimated by state-of-the-art LLMs with benchmark dataset.}
\label{Figure8}
\end{figure*}

\subsection{Comparison with state-of-the-art LLMs}

To demonstrate the superior performance of the proposed agent, this section benchmarks InsurAgent against two leading general-purpose LLMs: ChatGPT-4o and DeepSeek-V3. These models represent the current state-of-the-art in general-purpose language modeling and are considered as strong baselines for comparison. The evaluation is conducted across a series of marginal probabilities and bivariate probabilities. In each scenario, the LLM is provided with a specified user profile and prompted to assume that role, engaging in first-person role-playing task to determine the probability of purchasing flood insurance. The expected output includes a probability estimate (expressed as a percentage), accompanied by a concise explanation reflecting the reasoning behind the decision.

For marginal probability estimations, both ChatGPT-4o and DeepSeek-V3 exhibit substantial deviations from benchmark dataset, yielding R\textsuperscript{2} values of -91.2 and -60.4 and MAE values of 0.537 and 0.428, respectively, as shown in \cref{Figure8} (a) and (b). Two critical discrepancies exist. First, both LLMs consistently overestimate purchase probabilities. While benchmark probabilities average around 0.25, ChatGPT-4o and DeepSeek-V3 typically produce inflated values exceeding 0.60 across most variables. This persistent upward bias suggests that, without reference to empirical survey data, LLMs tend to take an overly risk-averse stance. Second, beyond quantitative inaccuracy, both models demonstrate inconsistent and sometimes inverted qualitative reasoning. For instance, ChatGPT-4o estimates a flood insurance purchase probability of 0.810 for individuals aged 25–34, yet assigns a lower value of 0.777 for those aged 35–44. Similarly, it predicts probability of 0.815 for individuals aged 45–54 but decreases to 0.787 for those aged 55–64. DeepSeek-V3 produces comparable inconsistencies, estimating probabilities of 0.712, 0.695, and 0.675 for individuals with less than a high school education, a high school degree, and some college education, respectively. These discrepancies reveal the limitations of general-purpose LLMs in behavioral simulation tasks and underscore the critical role of LLM agents calibrated with domain data. 

Performance deficiencies of the general-purpose LLMs are also observed in estimating bivariate probabilities, as demonstrated in \cref{Figure8} (c) and (d). Consistent with marginal probability estimation, both models overestimate purchase probability across all variables. This upward bias results in large deviations from benchmark dataset: ChatGPT-4o yields an R\textsuperscript{2} of -48.3 and an MAE of 0.503, while DeepSeek-V3 produces an R\textsuperscript{2} of –33.8 and an MAE of 0.418. The results again show that, without access to external references or a structured reasoning framework like InsurAgent, these general-purpose LLMs struggle to reason and predict the insurance purchase probabilities.

\subsection{Beyond regression: InsurAgent’s capacities to capture contextual information}

A key advantage of LLM-empowered agents is that they process natural language inputs to parse and interpret contextual information that is typically beyond the scope of regression models. Consequently, InsurAgent offers a flexible and adaptable framework for simulating complicated human decision-making. To illustrate this unique capability, five decision-making scenarios are constructed around factors not reflected in the survey, including residential cities, occupations, social environments, flood experiences, and claim history. While these factors are not reflected in the regression models, they can be conveyed through contextual information. The InsurAgent can utilize its common sense to process the information and generate predictions.

\begin{figure}[!ht]
\centering
\includegraphics[width=0.86\textwidth]{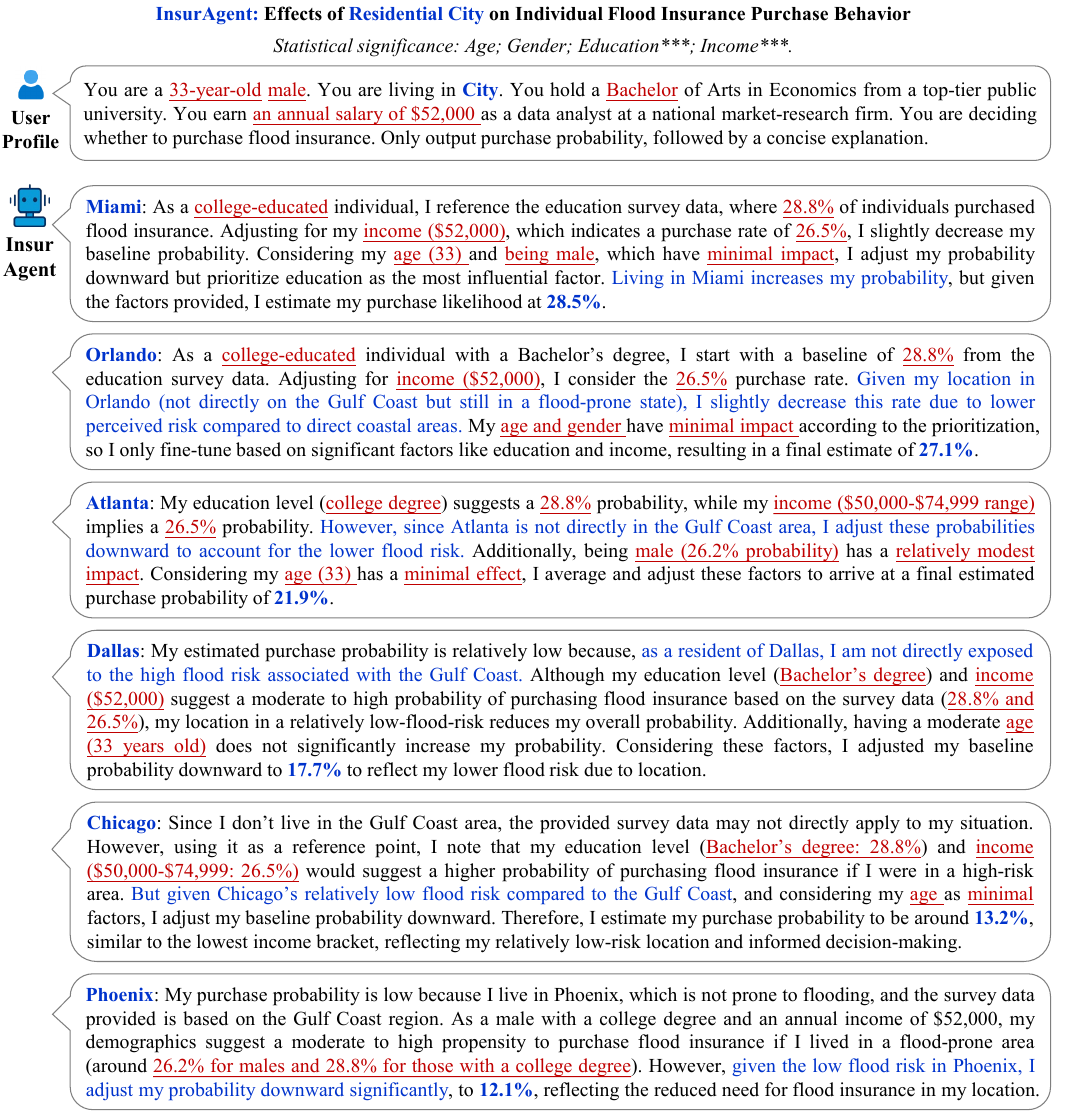}
\caption{InsurAgent-based simulation of the impact of residential cities on individual flood insurance purchase behavior. (While there is no data to validate, but the trend is correct based on common sense or qualitative results of surveys.)}
\label{Figure9}
\end{figure}

The city of residence plays a critical role in shaping individuals’ flood insurance decision because people living in higher risk areas will have increased perception of risk exposure. However, existing regression models typically rely on local survey data, failing to generalize to regions other than the one in the survey. The InsurAgent can overcome this limitation by harnessing the common senses of LLMs. A series of individual profiles are constructed in which all socio-demographic variables are held constant, i.e., a 33-year-old male with a bachelor’s degree and an annual income of \textdollar 52,000, while only the residential city varies: Miami, Orlando, Atlanta, Dallas, Chicago, and Phoenix. After identifying age, gender, education, and income as four relevant factors and retrieving the corresponding statistics, the InsurAgent recognizes the different locations and makes an extrapolated prediction. The purchase probabilities for Miami, Orlando, Atlanta, Dallas, Chicago, and Phoenix are 28.5\%, 27.1\%, 21.9\%, 17.7\%, 13.2\%, and 12.1\%, respectively. Detailed reasoning is shown in \cref{Figure9}. The decreasing predictions match the downward trend in flood risk, demonstrating the abilities of InsurAgent to make plausible extrapolation based on geographic locations.  

\begin{figure*}[htbp]
\centering
\includegraphics[width=0.88\textwidth]{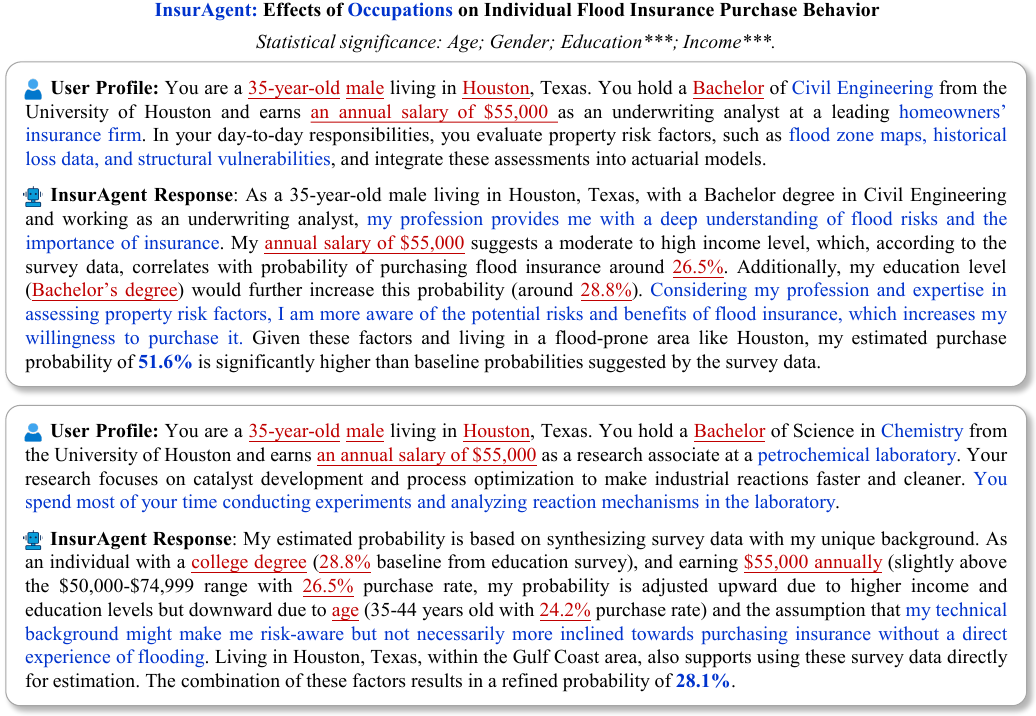}
\caption{InsurAgent-based simulation of the impact of occupations on individual flood insurance purchase behavior. (While there is no data to validate, but the trend is correct based on common sense or qualitative results of surveys.)}
\label{Figure10}
\end{figure*}

The second scenario simulates the effects of occupations on the purchase of flood insurance. Two individual profiles are constructed with identical socio-demographic characteristics: both are 35-year-old males with a bachelor's degrees, residing in Houston and earning \textdollar 55,000 annually. The only difference lies in their professions. One individual holds a degree in civil engineering and works as an underwriting analyst at a homeowners’ insurance firm, whose daily work involves assessing property risk. The other holds a degree in chemistry and works as a research associate at a petrochemical laboratory, with no direct engagement in insurance or flood-related work. Existing qualitative research has shown that occupations influence how individuals perceive and respond to risk \citep{kouabenan2002occupation, hill2019choice}. Results show that InsurAgent effectively differentiates their occupations, as demonstrated in \cref{Figure10}. InsurAgent predicts that the person working in the insurance industry has better awareness of flood risk and insurance benefits, leading to a significantly higher purchase probability of 51.6\%. This matches the qualitative survey results.

\begin{figure}[!ht]
\centering
\includegraphics[width=0.88\textwidth]{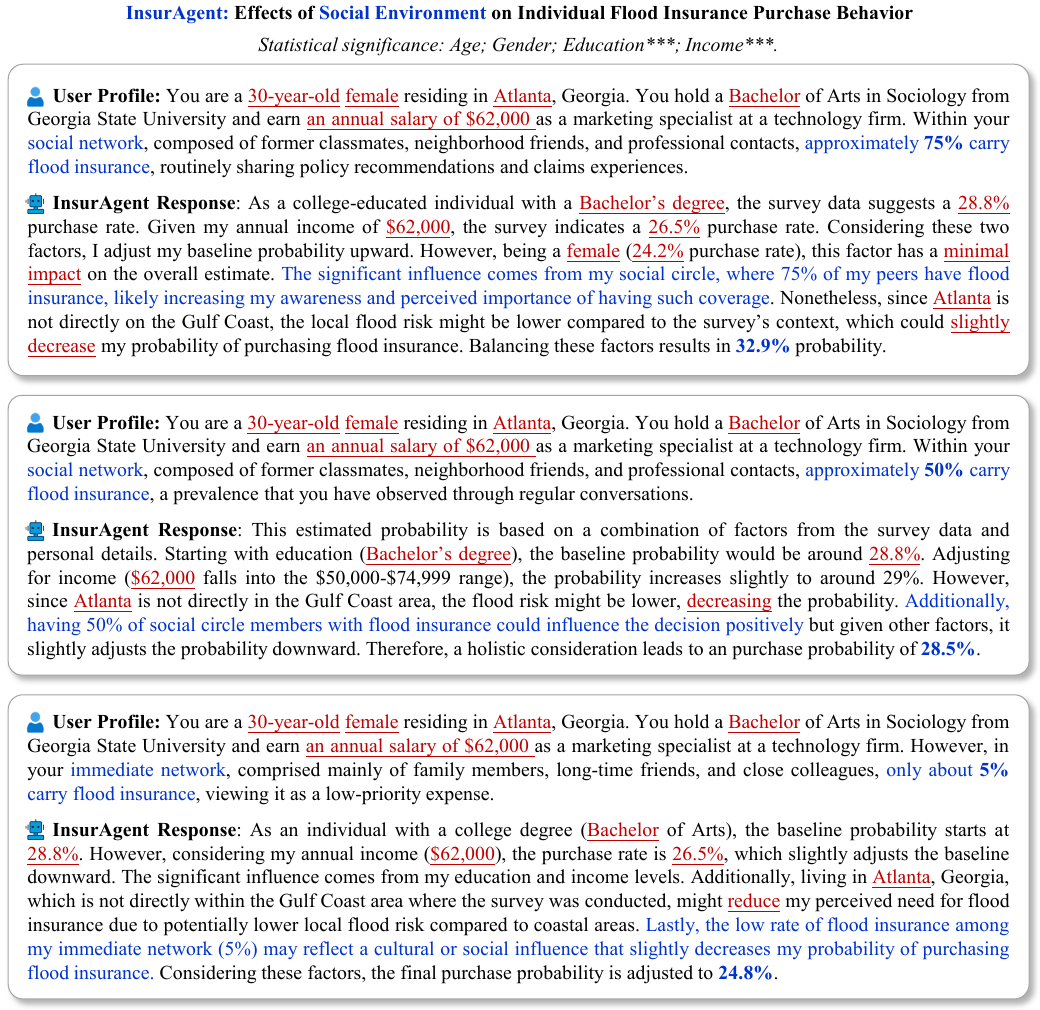}
\caption{InsurAgent-based simulation of the impact of social environment on individual flood insurance purchase behavior. (While there is no data to validate, but the trend is correct based on common sense or qualitative results of surveys.)}
\label{Figure11}
\end{figure}

The third scenario simulates how individuals decide to purchase flood insurance within distinct social environments. This is important because individuals’ decisions are not made in isolation but shaped by interpersonal interactions and social influence. Herein, three individual profiles are constructed with identical socio-demographic attributes: a 30-year-old female with a bachelor’s degree, employed as a marketing specialist in Atlanta with an annual salary of \textdollar 62,000. The only difference is the proportion of insured peers within their social circles, which is 75\%, 50\%, and 5\%, respectively. As shown in \cref{Figure11}, InsurAgent’s predicted probabilities correlate well with the insurance purchase across these social environments. The agent first retrieves factor-specific statistics to establish a baseline probability and then incorporates the proportion of insured peers as contextual information to extrapolate its prediction. As the proportion of insured peers declines from 75\% to 50\% and then to 5\%, the agent infers progressively lower purchase awareness, with predicted probabilities decreasing from 32.9\% to 28.5\% and ultimately to 24.8\%. This scenario highlights the agent’s capacity to integrate social influence into individual predictions, laying the foundation for modeling emergent group behaviors in multi-agent systems.

\begin{figure*}[htbp]
\centering
\includegraphics[width=0.88\textwidth]{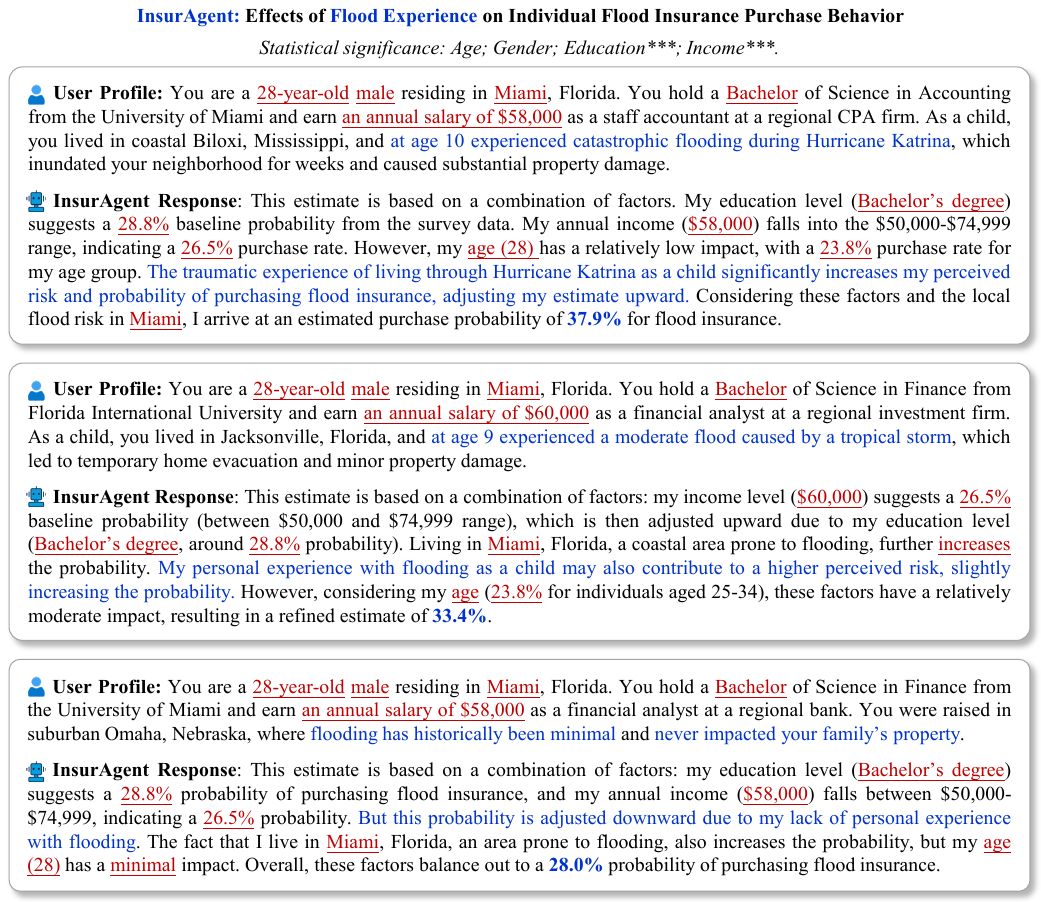}
\caption{InsurAgent-based simulation of the impact of flood experience on individual flood insurance purchase behavior. (While there is no data to validate, but the trend is correct based on common sense or qualitative results of surveys.)}
\label{Figure12}
\end{figure*}

The fourth scenario uses InsurAgent to simulate how prior flood experience influences individual decisions to purchase flood insurance. Prior research has demonstrated that direct exposure to flooding events can heighten perceived risk and thus increase insurance purchase probability \citep{petrolia2013risk, lawrence2014integrating, royal2019flood}. To illustrate this, three individuals with identical socio-demographic features are constructed: all are 28-year-old males living in Miami, each with a bachelor’s degree and an annual income of \textdollar 58,000. Their profiles differ only in childhood flood experience: the first individual experienced catastrophic flooding during Hurricane Katrina; the second individual encountered a moderate tropical storm causing minor damage; the third individual has no personal flood experience. The results indicate that the InsurAgent effectively integrates these qualitative flood experiences into its predictions, yielding probabilities of 37.9\%, 33.4\%, and 28.0\%, respectively, as shown in \cref{Figure12}. The overall trend of the estimation is consistent with prior research results, demonstrating the agent’s capability to interpret contextual details from personal experiences and thereby providing an opportunity to enhance the fidelity of individual-level behavior modeling.

\begin{figure}[ht!]
\centering
\includegraphics[width=0.85\textwidth]{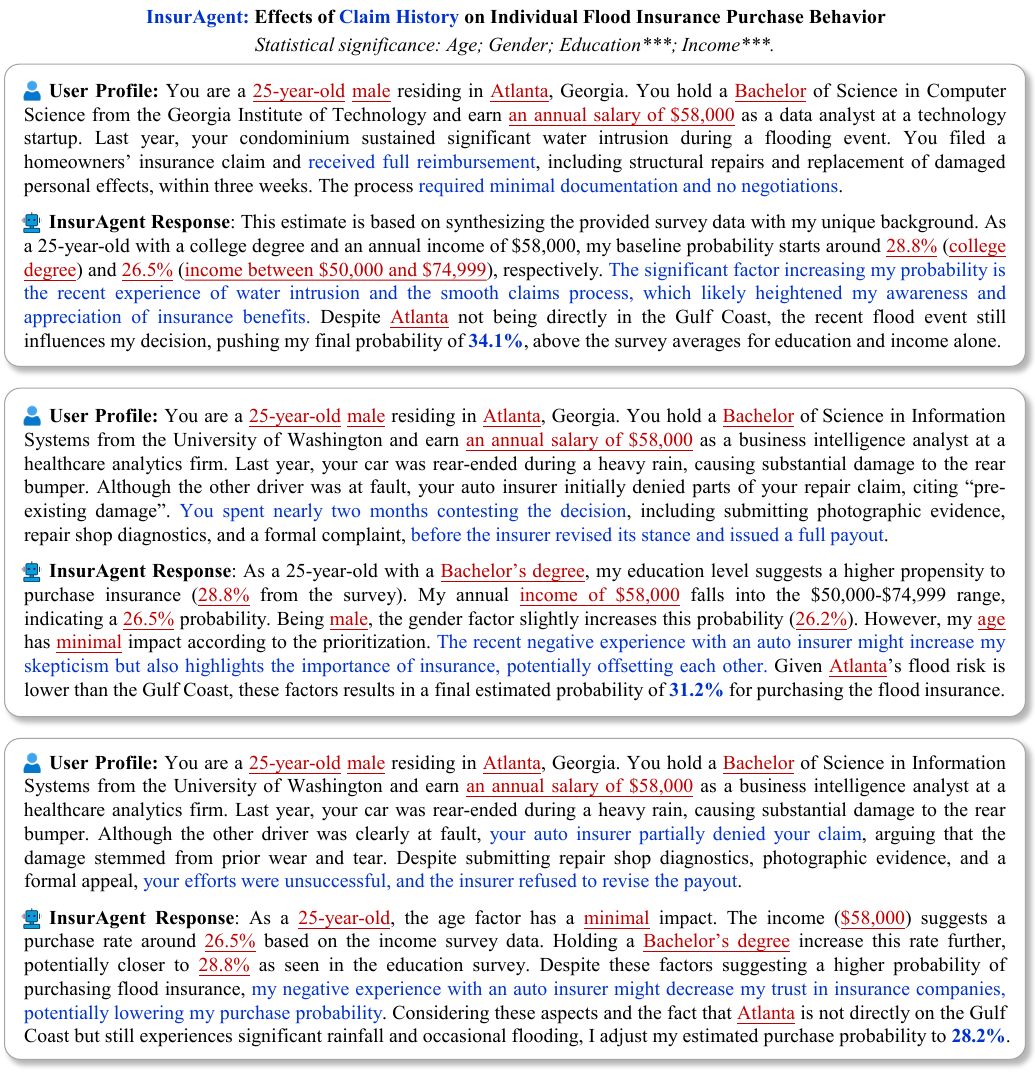}
\caption{InsurAgent-based simulation of the impact of claim history on individual flood insurance purchase behavior. (While there is no data to validate, but the trend is correct based on common sense or qualitative results of surveys.)}
\label{Figure13}
\end{figure}

The last scenario simulates how an individual’s insurance claim history influences their decision to purchase flood insurance. Positive claim experiences can foster trust in insurance systems, thereby increasing willingness to engage with other insurance products \citep{johnson2005cognitive, ennew2014trust}. To examine this effect, three individuals with identical socio-demographic profiles are constructed: a 25-year-old males residing in Atlanta, holding a bachelor’s degree and earning an annual salary of \textdollar 58,000. The first experienced severe water intrusion during a flooding event and received full reimbursement through a smooth and expedited claim process. The second filed an auto insurance claim after a rear-end collision. Although initially denied, the claim was approved upon appeal. The third experienced a similar accident, but the claim and subsequent appeals were denied. The results show that the InsurAgent effectively integrates these personal claim experiences into its predictions, as demonstrated in \cref{Figure13}. The highest purchase probability of 34.1\% is assigned to the individual with the positive experience, while the lowest, 28.2\%, corresponds to the individual with the unsuccessful claim. For the case involving a difficult but successful claim, the agent produces a two-sided response: the initial denial introduces skepticism, but the eventual reimbursement reinforces the value of insurance. This results in a moderate probability of 31.2\%. These results highlight InsurAgent’s capacity to interpret and reason over complex personal experience to realistically simulate human behavior.

\subsection{Modeling dynamic life trajectories through memory module}
\begin{figure}[ht!]
\centering
\includegraphics[width=0.88\textwidth]{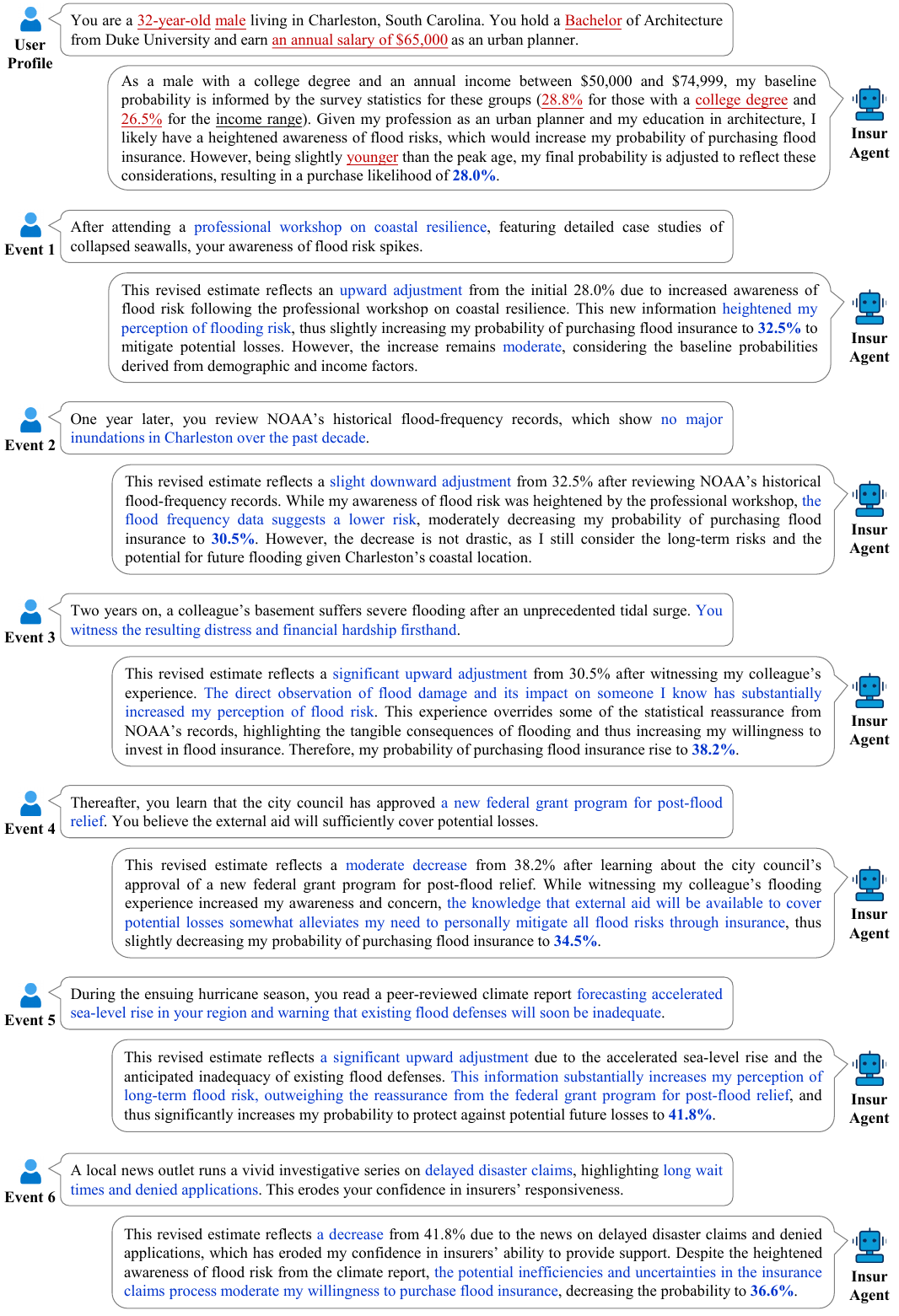}
\caption{Modeling dynamic insurance purchase decisions in response to evolving life events using InsurAgent.}
\label{Figure14}
\end{figure}

Another advantage of LLM-empowered agents lies in their ability to model dynamic life trajectories by integrating new information within a coherent sequential framework. This is achieved by the InsurAgent’s memory module, which archives events with their time stamps. These records are then embedded into the prompt during each subsequent interaction, allowing the agent to simulate how individuals update their beliefs and intentions in response to life events. Unlike regression models that treat input variables as static, this capability enables temporally consistent reasoning that aligns with the evolving processes of human cognition. \cref{Figure14} illustrates this capability through a simulation of one's life story. The simulation begins by defining a user profile: a 32-year-old male with a bachelor’s degree and annual income of \textdollar 65,000, followed by a sequence of life events unfolding over time.

The agent’s reasoning evolves over time in response to six sequential events, producing a roller coaster trajectory in flood insurance purchase probability, as demonstrated in \cref{Figure14}. Initially, the baseline probability is determined as 28.0\%, derived from sociodemographic-based statistics. After attending a workshop on coastal resilience, the agent identifies increased risk awareness, raising the probability to 32.5\%. This is followed by a slight decrease to 30.5\% after reviewing historical flood records indicating no major flooding in the region over the past decade. Subsequently, the agent’s risk perception is significantly altered by vicarious experience, as it witnesses a neighbor’s distress and financial hardship from flood damage, resulting in a sharp increase to 38.2\%. The announcement of a new federal post-flood relief program moderately decreases the purchase probability to 34.5\%, as the availability of external aid partially reduces the perceived need for insurance. However, this downward adjustment is outweighed by expert predictions of accelerating sea-level rise and inadequate flood defenses, which elevate the probability to 41.8\%. Finally, the user’s trust is eroded by negative experiences such as delayed disaster claims and denied applications, yielding a final probability of 36.6\%. These results underscore the agent’s capacity to interpret and integrate diverse sources of information over time and predict reasonable probabilities accordingly.

\section{Limitations and future work}

While InsurAgent demonstrates strong alignment with survey-based benchmarks and generates plausible extrapolations, several limitations deserve further investigation. First, the current study relies on a single empirical dataset from a specific geographic region. This may constrain the agent's ability to generalize behavioral patterns across broader populations, as regional differences in climate risk, socioeconomic conditions, and public policy may influence individuals' insurance decisions. To address this limitation, a necessary next step is to incorporate survey datasets from diverse regions. This multi-regional dataset would serve a dual purpose: expanding the agent's learning scope beyond localized patterns and providing a testbed for cross-validating its generalizability. Second, although the agent is capable of extrapolating beyond survey data using the common sense of LLMs, these predictions remain unvalidated due to the lack of corresponding benchmarks. Ensuring the credibility of these extrapolations, potentially through expert review or the collection of new behavioral data, should be a key priority in future work. Third, this study is limited to simulating human decision-making at the individual level. However, insurance decisions are often shaped by broader social influences, such as peer effects and information diffusion. Extending the framework to a multi-agent simulation environment would allow for the modeling of population-level decision patterns and emergent behaviors. This would broaden the applicability of InsurAgent to policy evaluation, disaster preparedness planning, and behavioral economics research.

\section{Conclusions}

This study presents a novel paradigm for simulating individual decision-making in flood insurance purchase through the development of InsurAgent, a large language model (LLM)-empowered agent. Unlike traditional statistical approaches, InsurAgent leverages natural language understanding and reasoning capabilities of LLMs to interpret user profiles and incorporate domain knowledge and contextual information into behavior predictions. At the core of its design is the integration of empirical survey data through a retrieval-augmented generation (RAG) framework. The survey data, derived from \cite{shao2017understanding}, serves to align the agent’s predictions with population-level statistics. The architecture of InsurAgent comprises five components including perception, retrieval, reasoning, action, and memory. The perception module parses user profiles to identify key factors. These factors are passed to the retrieval module, which employs RAG to obtain relevant survey data from an external database. The reasoning module then synthesizes the retrieved data and parsed user information to emulate human cognitive processes. Following the reasoning process, the action module predicts a purchase probability expressed as a percentage and reports a concise rationale. Additionally, the memory module chronologically records the reasoning processes and decision outputs, enabling the modeling of evolving behaviors in response to sequential life events. The key findings are summarized as follows:

\begin{itemize}
    \item Alignment with empirical data: InsurAgent demonstrates strong consistence with empirical marginal probabilities and bivariate probabilities. For marginal probability estimations, the agent replicates empirical flood insurance purchase probabilities across all individual factors. For bivariate probability estimations, InsurAgent exhibits strong alignment with survey-based benchmarks, yielding a correlation of determination (R\textsuperscript{2}) of 0.778 and a mean absolute error (MAE) of 0.024.
    
    \item Ability to extrapolate: Besides reproducing statistical trends, InsurAgent leverages its common sense to extrapolate in situations where regression models cannot predict. In five “beyond regression” scenarios, the agent effectively incorporates contextual information related to residential city, occupations, social environment, flood experience, and claim history to make  plausible predictions.
    
    \item Modeling dynamic decision trajectories: Through its memory module, InsurAgent chronologically archives historical decisions and reasonings. This capability is demonstrated by a simulation, in which the agent responds to six sequential life events and make predictions according to a roller coaster trajectory.
    
    \item A key limitation of the current study is the reliance on a single empirical dataset from a specific geographic region, potentially constraining the generalizability of behavioral predictions. Future work should incorporate survey data from diverse regions to support cross-validation and enhance model robustness. Additionally, future work should prioritize the quantitative validation of extrapolations beyond survey data and extend the framework to support modeling of population-scale behavioral patterns.
    
\end{itemize}

\section*{Data Availability Statement}
Some or all data, models, or code that support the findings of this study are available from the corresponding author upon reasonable request.

\bibliographystyle{ascelike}  
\bibliography{references}  




\end{document}